\RequirePackage{fix-cm}
\documentclass{article}
\usepackage[table]{xcolor}
\usepackage{hyperref}

\usepackage[preprint]{report}

\usepackage{microtype}
\usepackage{roboto}
\usepackage{graphicx}
\usepackage{subcaption}
\usepackage{booktabs}
\usepackage{multirow}
\usepackage{amssymb}
\usepackage{needspace}
\usepackage{placeins}
\usepackage{tcolorbox}
\tcbuselibrary{skins}
\usepackage{amsmath,amsfonts,bm}

\def\eqref#1{equation~\ref{#1}}
\def\Eqref#1{Equation~\ref{#1}}

\def\1{\bm{1}}

\DeclareMathAlphabet{\mathsfit}{\encodingdefault}{\sfdefault}{m}{sl}
\SetMathAlphabet{\mathsfit}{bold}{\encodingdefault}{\sfdefault}{bx}{n}

\definecolor{ourblue}{HTML}{E3EEF9}

\newcommand{\h}{\mathbf{h}}

\newcommand{\rms}{\operatorname{RMSNorm}}
\newcommand{\method}{InfiLoop}
\newcommand{\methodshort}{\method}

\icmltitlerunning{Loop-Native Attention Residuals}
\hypersetup{pdfauthor={Pengxiang Li, Dilxat Muhtar, Di He, Guinan Su, Lu Yin, Shiwei Liu},pdftitle={Scaling to Tens of Thousands of Test-Time Iterations with Loop-Native Attention Residuals}}

\definecolor{arxivcream}{RGB}{254,248,242}
\definecolor{arxivaffil}{RGB}{82,76,71}
\makeatletter
\renewcommand{\printAffiliationsAndNotice}[1]{\global\icml@noticeprintedtrue}
\makeatother

\begin{document}

\thispagestyle{plain}
\begin{tcolorbox}[
  enhanced,
  colback=arxivcream,
  colframe=arxivcream,
  boxrule=0pt,
  arc=7pt,
  left=16pt, right=16pt, top=14pt, bottom=12pt,
  before skip=0pt, after skip=12pt,
  width=\linewidth
]
\begin{center}
\setlength{\parskip}{0pt}%
{\robotoblack\fontsize{17.5}{19.4}\selectfont
Scaling to Tens of Thousands of Test-Time Iterations\\with Loop-Native Attention Residuals\par}
\vspace{14pt}
{\robotobold\fontsize{11}{12.6}\selectfont
Pengxiang Li\textsuperscript{1},
Dilxat Muhtar\textsuperscript{2},
Di He\textsuperscript{3,4,5},
Guinan Su\textsuperscript{6,7,8},
Lu Yin\textsuperscript{9},
Shiwei Liu\textsuperscript{6,7,8,\dag}\par}
\vspace{12pt}
{\robotoregular\fontsize{8.6}{12.8}\selectfont\color{arxivaffil}
\textsuperscript{1}The Hong Kong Polytechnic University,
\textsuperscript{2}Independent Researcher,\par
\textsuperscript{3}Shenzhen Institutes of Advanced Technology, Chinese Academy of Sciences,\par
\textsuperscript{4}Peng Cheng Laboratory,
\textsuperscript{5}University of Chinese Academy of Sciences,\par
\textsuperscript{6}ELLIS Institute T\"ubingen,
\textsuperscript{7}Max Planck Institute for Intelligent Systems,\par
\textsuperscript{8}T\"ubingen AI Center,
\textsuperscript{9}Shenzhen University of Advanced Technology\par
\textsuperscript{\dag}Corresponding author: sliu@tue.ellis.eu\par}
\end{center}
\vspace{10pt}
{\rmfamily\fontsize{10.2}{14.6}\selectfont
\setlength{\parindent}{0pt}%
In this paper, we argue that Looped Transformers need their own residual connections to prevent performance degradation as the number of iterations grows. We first introduce an interesting phenomenon we observed in Looped Transformers, called \emph{latent overthinking}: noisy state updates overwrite correct intermediate deductions and even undo completed solutions. This leaves subsequent iterations to recover lost information from an already degraded representation: once an error arises in an earlier loop, often as a result of long-range propagation through the recurrence, later loops find it difficult to correct. In this paper, we introduce \texttt{InfiLoop}, a loop-native residual connection that learns which past computations to retain and how much to accept from each new update. InfiLoop combines content-based weighting with learned temporal decay to maintain a running summary of recurrent states. An exact streaming recurrence keeps its persistent aggregation memory constant as the loop count grows. The resulting adaptive update suppresses unreliable proposals and preserves useful intermediate states. Across extensive reasoning tasks, a 7M-parameter InfiLoop model outperforms existing recursive architectures, reaching 97.9\% exact accuracy on Sudoku-Extreme, and 13.6\% pass@2 on ARC-AGI-2. Notably, on Sudoku-Extreme, \texttt{InfiLoop} continues to improve with test-time looping beyond 20,000 effective steps, showing that added depth translates directly into stronger reasoning. Our code is available at \url{https://github.com/pixeli99/InfiLoop}.\par}
\end{tcolorbox}
\printAffiliationsAndNotice{}

\section{Introduction}
\label{sec:introduction}
Looped Transformers address complex reasoning tasks by repeatedly applying a shared Transformer block~\citep{dehghani2019universal,wang2025hierarchical,jolicoeurmartineau2025less}.
Each iteration adds another sequence of layer transformations, increasing computational depth without introducing additional parameters.
This allows models to devote more computation to reasoning by executing more loops at test time~\citep{saunshi2025reasoning,geiping2025scaling}.
A common way to connect consecutive iterations is to feed the final output of one loop directly into the next, a strategy we refer to as the \emph{carry-last rule}.
Although effective, this rule makes the unrolled computation equivalent to a deep network with weights shared across repeated blocks.
Increasing the loop count therefore returns us to a familiar question: \emph{does adding more depth lead to more useful computation?}

Standard deep Transformers suffer from the \emph{curse of depth}~\citep{sun2025curse}, where the contribution of deeper layers progressively diminishes.
Repeatedly applying a shared block can compound this limitation, as the same transformations are applied at every iteration and the residual stream grows with the loop count. 
Related limitations have also been examined in fixed-point looping~\citep{movahedi2026fixed}.
Existing remedies use decay or layer scaling to control residual growth and regulate update magnitudes~\citep{sun2025curse,wang2026residual,movahedi2026fixed}.
These interventions address how strongly each update affects the representation, but leave a central architectural choice unchanged: the next loop still receives only the latest recurrent state.

We argue that this state-replacement rule is a key bottleneck in deep looped transformers.
Under carry-last, earlier computations remain accessible only through the information preserved in the latest state.
Standard Pre-LN or Post-LN residual connections provide paths for this information to propagate, but do not explicitly retain and select among outputs from earlier loops.
Consequently, a noisy or undesirable iteration can degrade useful representations throughout propagation, leading to sub-optimal performance, as shown in Figure \ref{fig:loop_progress}. We call this phenomenon \emph{latent overthinking}: additional loops lower accuracy because later updates overwrite useful intermediate states in the latent recurrent state. The term follows the overthinking that \citet{bansal2022overthinking} observed in recurrent networks, here arising in the latent state of looped Transformers. This motivates changing how information is combined across loops, so that further computation can build on useful intermediate states without relying exclusively on the most recent output.

To this end, we introduce \textbf{\method{}}, a loop-native residual connection that replaces the carry-last rule with learned attention over recurrent outputs.
A shared pseudo-query scores each candidate when it is produced, while a learned exponential decay discounts earlier contributions.
The resulting normalized aggregation forms the input to the next iteration.
Crucially, shared scoring and geometric decay admit an exact streaming recurrence, requiring only a weighted-state vector and a normalization scalar per token, with persistent aggregation memory independent of the loop count.
This recurrence is equivalent to an adaptive residual update that balances the accumulated state against each new candidate, allowing the model to regulate how strongly subsequent computations alter its reasoning state.

This paper establishes four findings.
First, empirical diagnosis reveals that carry-last destabilizes solved states at large depth, while \methodshort{} preserves solutions indefinitely.
Second, the exact recurrence guarantees constant-memory state combination across arbitrarily long reasoning trajectories.
Third, on extensive reasoning benchmarks, \methodshort{} surpasses strong damping baselines while requiring substantially fewer layer evaluations.
Fourth, inspection of the learned step size reveals an emergent self-verification behavior: the connection discounts outputs produced during unsolved search, accepts the output that completes the solution, and freezes the state once solved.

\section{Latent Overthinking: Deeper Loops Undo Earlier Progress}
\label{sec:diagnosis}

\begin{figure}[!t]
\centering
\includegraphics[width=1.0\linewidth]{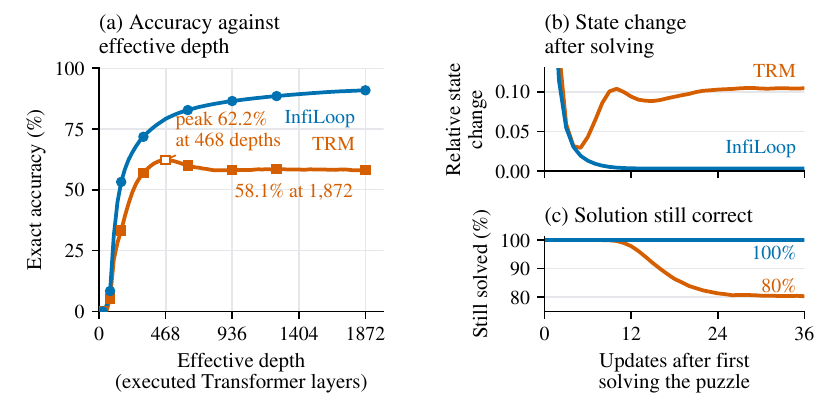}
\caption{\textbf{Carry-last keeps changing its state after solving and loses solutions.}
(a) Exact accuracy against effective depth, the number of executed Transformer layers, on Sudoku-Extreme.
Each outer update executes 26 layers.
TRM peaks at 468 layers and then loses accuracy as depth grows.
(b) Relative state change after the update at which each model first solves a puzzle, for the puzzles that both models solve within 36 updates.
TRM settles briefly and then moves its state again.
(c) Share of those puzzles whose solution is still correct.}
\label{fig:loop_progress}
\end{figure}

We measure where a looped reasoner breaks as depth grows.
On Sudoku-Extreme, we compare \methodshort{} with the self-attention TRM, a carry-last model.
Both checkpoints are trained with the same loop schedule, and we run both for 72 outer updates.
After each outer update, we decode the predicted board and measure the relative change of the outer state, $\|\h_t-\h_{t-1}\|/\|\h_{t-1}\|$.
We average this change over initially empty cells.
Figure~\ref{fig:loop_progress} plots exact accuracy against effective depth, the number of executed Transformer layers, and tracks the state change after solving.

\paragraph{Accuracy peaks and then falls with depth.}
TRM reaches its best exact accuracy of 62.2\% at 468 layers, but its accuracy declines to 58.1\% at 1,872 layers.
\methodshort{} keeps improving over the same depth and reaches 90.9\%.
The predictions show the same pattern.
During updates 36 to 72, TRM changes 77.0 predictions per puzzle on initially empty cells, and its number of incorrect cells rises by 0.21 per puzzle.
Repeated changes to the same cell count separately.
Over the same window, \methodshort{} makes 24.0 changes and removes 1.25 incorrect cells per puzzle.

\paragraph{Correct intermediate predictions are lost.}
TRM overwrites correct predictions in later updates.
At update 36, both models predict 1.42 million initially empty cells correctly.
During the remaining updates, TRM turns 20.9\% of these predictions into an incorrect value at least once, and \methodshort{} turns 2.4\% of them.
The right two panels of Figure~\ref{fig:loop_progress} align each puzzle to its first solving update.
Within 5 updates after solving, TRM reduces its state change to 3\% of the state norm, but the change then climbs back to 10\% per update and stays there.
TRM begins to lose solutions after 9 updates and has lost 19.7\% of them after 36 updates.
The state change of \methodshort{} instead falls to 0.3\% per update, and every puzzle solved by \methodshort{} remains solved 36 updates later.

\paragraph{Learning what to carry forward.}
A newer output is not necessarily a better output, yet carry-last passes every new block output to the next iteration in full.
Earlier outputs reach later iterations only through the newest output, so the connection between iterations decides what the loop keeps.
We therefore replace carry-last with a learned residual connection.
A running summary holds the contributions of earlier outputs, and the connection learns how much of each new output to add to it.
The summary becomes the input to the next iteration, so the learned connection acts during reasoning and not on predictions after reasoning ends.
A shared scorer weights each output once, at its own iteration, and a learned decay reduces the contributions of older outputs.
The weighted summary has an exact recurrent form, and its persistent memory does not grow with the number of iterations.
Section~\ref{sec:method} defines the connection and derives its recurrence.

\FloatBarrier
\section{Loop-Native Attention Residuals}
\label{sec:method}

\begin{figure}[!t]
\centering
\includegraphics[width=0.95\linewidth]{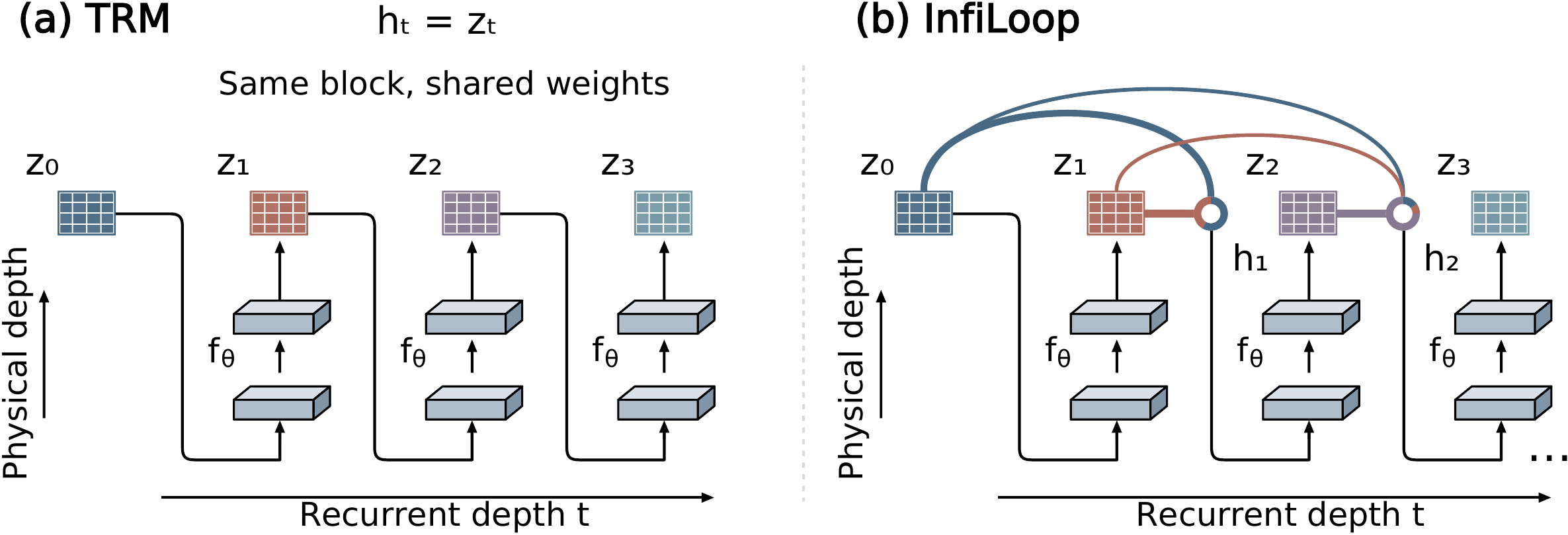}
\caption{\textbf{Connecting successive applications of a shared block.}
(a) TRM uses the carry-last rule: the candidate $z_t$ becomes the next recurrent state, $h_t=z_t$.
(b) \methodshort{} combines earlier candidates into the state $h_t$ passed to the next block application, using content weights and exponential temporal decay.
Colors identify candidates from different iterations; the illustrated contributions are schematic.
The same block parameters are reused at every recurrent step.}
\label{fig:overview}
\end{figure}

A looped Transformer applies a shared block $f_\theta$ for $T$ recurrent steps, conditioned on an input representation $x$.
Under the commonly used \emph{carry-last} rule, each step produces a candidate state $z_t$ and passes it directly to the next iteration:
\begin{equation}
z_t = f_\theta(h_{t-1}; x), \qquad h_t = z_t,
\label{eq:carrylast}
\end{equation}
where $h_0$ is the initial recurrent state.
Earlier computations remain accessible only through the information retained in the latest state.
Figure~\ref{fig:overview} contrasts this connection with the learned aggregation used by \methodshort{}.

\subsection{Attention Residuals Across Recurrent Steps}
\label{sec:formulation}

Attention Residuals constructs the residual stream by attending over representations from earlier layers~\citep{kimi2026attention}.
We adapt this principle to recurrent steps, combining content-based weighting with temporal decay.
The aggregation operates independently at each token position; below, we omit the token index and write $z_i,h_t \in \mathbb{R}^d$, where $d$ is the hidden dimension.
Let $z_0=h_0$ denote the initial representation and $z_i$ the candidate produced at step $i\geq 1$.
The state passed to the next iteration is
\begin{equation}
h_t = \sum_{i=0}^t \alpha_{t,i} z_i,
\qquad
\alpha_{t,i}
=
\frac{\exp(s_i-\lambda(t-i))}
{\sum_{j=0}^t \exp(s_j-\lambda(t-j))}.
\label{eq:attn_res}
\end{equation}
Here $\alpha_{t,i}$ is the normalized weight assigned to candidate $z_i$ at step $t$.
Its content score is $s_i=q^\top\rms(z_i)$, where $q\in\mathbb{R}^d$ is a learned pseudo-query shared across iterations and $\rms(\cdot)$ denotes RMS normalization \citep{zhang2019rmsnorm}.
Each candidate is therefore scored once, when it is produced.
The temporal bias $-\lambda(t-i)$ discounts older candidates using a learned decay rate $\lambda>0$.
The unnormalized weight factors as $\exp(s_i)\exp(-\lambda(t-i))$, combining content-based relevance with recency.
In the no-decay case $\lambda=0$, the aggregation recovers the Full Attention Residuals rule over recurrent outputs, with the pseudo-query tied across iterations.

Equivalently, the aggregated state can be expressed as the ratio of an unnormalized weighted sum $N_t\in\mathbb{R}^d$ to a normalization mass $D_t\in\mathbb{R}_{>0}$:
\begin{equation}
N_t = \sum_{i=0}^t \exp(s_i-\lambda(t-i))z_i,
\qquad
D_t = \sum_{i=0}^t \exp(s_i-\lambda(t-i)),
\qquad
h_t = \frac{N_t}{D_t}.
\label{eq:accumulators}
\end{equation}
These quantities admit an exact streaming update, derived below.

\subsection{Exact Recurrence with Constant Memory}
\label{sec:recurrence}

Directly evaluating the sums in \Eqref{eq:accumulators} would require retaining the full candidate history.
The shared scorer and exponential temporal decay make this unnecessary: historical content scores remain unchanged, and their weights decay by the same factor at each step.
Specifically, let $e_i=\exp(s_i)$ and $\beta=e^{-\lambda}\in(0,1)$, so that
$\exp(s_i-\lambda(t-i))=e_i\beta^{t-i}$.
Separating the newest contribution gives
\begin{equation}
\begin{aligned}
N_t
&= \sum_{i=0}^{t-1}\beta^{t-i}e_i z_i + e_tz_t
 = \beta N_{t-1}+e_tz_t,\\
D_t
&= \sum_{i=0}^{t-1}\beta^{t-i}e_i + e_t
 = \beta D_{t-1}+e_t.
\end{aligned}
\label{eq:recurrence}
\end{equation}
We initialize $N_0=e_0z_0$ and $D_0=e_0$, which yields $h_0=z_0$, and recover the state at each step as $h_t=N_t/D_t$.
The recurrence is algebraically identical to \Eqref{eq:attn_res}; no approximation to the attention weights is introduced.

\subsection{Interpretation as an Adaptive Residual Update}
\label{sec:interpretation}

Substituting $N_{t-1}=D_{t-1}h_{t-1}$ into \Eqref{eq:recurrence} and dividing by $D_t$ yields
\begin{equation}
h_t = h_{t-1}+\gamma_t(z_t-h_{t-1}), \qquad
\gamma_t = \frac{e_t}{\beta D_{t-1}+e_t}\in(0,1).
\label{eq:step_size}
\end{equation}
The attention aggregation thus implements an adaptive step toward the new candidate.
The step size $\gamma_t$ compares the candidate's content weight $e_t$ with the retained historical mass $\beta D_{t-1}$.
When $e_t\gg\beta D_{t-1}$, $\gamma_t$ approaches one and the new candidate dominates.
When $e_t\ll\beta D_{t-1}$, $\gamma_t$ approaches zero and the previous state is largely preserved.
The learned scorer can therefore suppress low-scoring proposals while accepting high-scoring ones. Because the attention weights are positive and sum to one, $h_t$ lies in the convex hull of the candidate states, i.e., $\|h_t\| \leq \max_{0\leq i\leq t}\|z_i\|$.
Thus, aggregation does not amplify the norm beyond that of the largest candidate.
A uniform bound over arbitrarily many iterations additionally requires the candidate norms to remain bounded.

Moreover, our framework applies to both standard looped Transformers \citep{dehghani2019universal} and nested recurrent reasoners like TRM~\citep{jolicoeurmartineau2025less} and HRM  \citep{wang2025hierarchical}.
For a standard looped Transformer, we replace the carry-last assignment with \Eqref{eq:recurrence}.
For nested looped architectures, we apply the same rule independently to each recurrent axis.
The inner and outer states each have a separate scorer--decay pair $(q,\beta)$ and accumulators $(N,D)$, adding $d+1$ learned parameters per axis while keeping the aggregation state independent of the loop count.

\subsection{Computational and Memory Complexity}

Our aggregation stores only a vector $N_t\in\mathbb{R}^d$ and a scalar $D_t$ per token, requiring $O(d)$ persistent memory and $O(d)$ computation per iteration.
Over $T$ iterations, its total aggregation cost is $O(Td)$, while its memory footprint remains independent of the loop count.
In contrast, Full Attention Residuals with iteration-specific queries stores and attends over all preceding representations, requiring $O(td)$ memory and computation at step $t$~\citep{kimi2026attention}.
Our recurrence therefore reduces peak persistent memory from $O(Td)$ to $O(d)$ and total aggregation cost from $O(T^2d)$ to $O(Td)$.

\FloatBarrier
\section{Experiments}
\label{sec:experiments}
\label{sec:setup}

We evaluate \methodshort{} against recursive and looped reasoning models across symbolic puzzle benchmarks and abstract visual transformation tasks.
Our implementation builds upon the public TRM codebase \citep{jolicoeurmartineau2025less} and adopts its deep-supervision training procedure.
We first describe the benchmarks and evaluation protocol, then report complete system performance, examine test-time compute and depth scaling, and analyze the emergent verification dynamics of the learned connection.

\subsection{Dataset description}
\label{sec:datasets}
In the following, we provide a structured description of each dataset used in this paper.
We refer the reader to the cited literature for further information.

\paragraph{Sudoku-Extreme.}
The task consists of exceptionally challenging, partially filled $9\times9$ Sudoku puzzles with unique solutions, introduced by \citet{wang2025hierarchical}.
Each board is flattened into a sequence of length 81.
The training set contains 1,000 unique base puzzles, each augmented 1,000 times through digit permutations and spatial symmetries, yielding approximately 1M training samples.
The test data contains 422,786 unique puzzles.
The evaluation metric is exact sequence accuracy, requiring all 81 board cells to match the ground truth.

\paragraph{Maze-Hard.}
The task consists of $30\times30$ shortest-path maze puzzles, introduced by \citet{wang2025hierarchical}.
Each grid contains open navigable cells, impassable walls, an origin, and a destination.
The model predicts a flattened grid marking the reference shortest path.
The training and test sets each contain 1,000 unique samples.
Following \citet{wang2025hierarchical}, we apply no data augmentation to this task.
The evaluation metric is exact sequence accuracy.

\paragraph{ARC-AGI-1 and ARC-AGI-2.}
The Abstraction and Reasoning Corpus assesses whether a reasoning system can induce an underlying transformation rule from minimal demonstrations \citep{chollet2019intelligence,chollet2025arcprize}.
ARC-AGI-1 provides two or more demonstration pairs (typically three) of coloured 2D grids with variable dimensions up to $30\times30$, and requires generating the output grid for a held-out test input.
ARC-AGI-2 keeps this format while presenting substantially more intricate problems that compose multiple transformations and resist brute-force search \citep{chollet2025arcagi2}.
ARC-AGI-1 has 400 public training tasks and 400 public evaluation tasks; ARC-AGI-2 has 1,000 public training tasks and 120 public evaluation tasks.\footnote{Official ARC-AGI-2 dataset: \url{https://github.com/arcprize/ARC-AGI-2}.}
The evaluation metric is pass@2 accuracy, selecting the top two candidate outputs aggregated across augmented test-time views.
Pretrained reasoning language models with chain-of-thought struggle on both benchmarks: DeepSeek-R1 with 671B parameters attains 15.8\% on ARC-AGI-1 and 1.3\% on ARC-AGI-2, while Claude 3.7 Sonnet (16K) attains 28.6\% and 0.7\%, respectively, as reported in Table~\ref{tab:main}.

\subsection{Evaluation protocol}
\label{sec:protocol}
\label{app:details}

\paragraph{Model architecture.}
We evaluate a compact 7M-parameter recursive reasoner that implements the nested looping architecture in Section~\ref{sec:interpretation}.
Its shared recursive block contains two Transformer layers with hidden dimension 512, eight attention heads, SwiGLU activations \citep{shazeer2020glu} with an expansion factor of four, and a depthwise $3\times3$ convolution.
Across both the inner and outer recurrent loops, the model replaces standard carry-last state replacement with our loop-native attention residual connection, parameterized by a 513-parameter scorer--decay pair $(q, \beta)$ per axis.

\paragraph{Baselines.}
We benchmark \methodshort{} against the primary looped reasoning paradigms:
(1) \textbf{TRM-Att} (7M parameters) \citep{jolicoeurmartineau2025less}, the standard self-attention recursive reasoner using carry-last state propagation;
(2) \textbf{FPRM} (7M parameters) \citep{movahedi2026fixed}, a released looped model that stabilizes iterations via pre-norm, learned residual scaling, and damped fixed-point updates;
(3) \textbf{HRM} \citep{wang2025hierarchical}, \textbf{Attractor Models} \citep{feinashley2026solve}, \textbf{EqR} \citep{huang2026equilibrium}, and \textbf{URM} \citep{gao2025universal}, spanning parameter budgets from 7M to 27M;
and (4) \textbf{Frontier reasoning LLMs} using test-time chain-of-thought.
For direct controlled comparisons, we evaluate the official released checkpoint of FPRM on Sudoku-Extreme and retrain the TRM baseline with the authors' published recipe.

\paragraph{Training configuration.}
For Sudoku-Extreme, we train \methodshort{} on 1,000 base puzzles with 1,000 augmentations per puzzle.
We optimize using AdamW \citep{loshchilov2019adamw} with peak learning rate $10^{-4}$, weight decay 1.0, and batch size 768.
The training schedule executes $H=3$ outer loops and $L=6$ inner updates across 12 deep-supervision segments.
Each segment executes 21 shared-block applications, with gradients backpropagated through the final seven.
For Maze-Hard and ARC benchmarks, training employs $H=3, L=6$ across up to 16 supervision segments with peak learning rate $3\times10^{-4}$.
Appendix~\ref{app:resources} provides complete hyperparameter and resource accounting.

\subsection{Reasoning performance across puzzle benchmarks}
\label{sec:benchmarks}

The complete \methodshort{} reasoner achieves the highest performance among the 7M-parameter configurations listed in Table~\ref{tab:main} across the four benchmarks.
On Sudoku-Extreme, \methodshort{} achieves 97.9\% exact accuracy, surpassing FPRM at 94.2\% and reported TRM-Att at 74.7\%.
On Maze-Hard, \methodshort{} achieves 87.4\% accuracy, exceeding both TRM-Att at 85.3\% and FPRM at 87.0\%.
On ARC-AGI-1, \methodshort{} reaches 49.1\% pass@2, improving upon FPRM at 47.5\% and TRM-Att at 44.6\%.
On the challenging ARC-AGI-2 benchmark, \methodshort{} attains 13.6\% pass@2, nearly doubling TRM-Att at 7.8\% and outperforming FPRM at 6.2\%.
We also reproduce the TRM baseline ourselves, marked $^\dagger$ in Table~\ref{tab:main}.
Retraining TRM on Sudoku-Extreme with the authors' published recipe gives 73.6\%, close to the 74.7\% they report.
Evaluating the released ARC-AGI-2 verification checkpoint under our protocol yields 4.6\% pass@2.
All baseline comparisons in the following sections use the retrained checkpoint.

\begin{table}[!tbp]
\centering
\caption{Reasoning benchmark results (\%) for complete systems under their respective inference procedures.
Sudoku and Maze use exact solution accuracy.
ARC uses pass@2.
Best results are bold, and the best among the listed 7M-parameter models are underlined.
Results for DeepSeek R1 \citep{deepseek2025r1}, Claude 3.7 Sonnet \citep{anthropic2025claude37}, o3-mini \citep{openai2025o3mini}, and Gemini 2.5 Pro \citep{comanici2025gemini25} follow \citet{jolicoeurmartineau2025less}; URM and EqR parameter counts follow \citet{movahedi2026fixed}.
$^\ddagger$URM reports ARC pass@1.
$^\dagger$Reproduced by us: TRM retrained on Sudoku-Extreme with the authors' published recipe, and the released ARC-AGI-2 verification checkpoint evaluated under our protocol.}
\label{tab:main}
\small
\setlength{\tabcolsep}{5pt}
\renewcommand{\arraystretch}{1.0}
\begin{tabular}{l c cc cc}
\toprule
& & \multicolumn{2}{c}{Puzzles (exact acc.)}
  & \multicolumn{2}{c}{ARC-AGI (pass@2)} \\
\cmidrule(lr){3-4}
\cmidrule(lr){5-6}
Method & Params & Sudoku-Ext. & Maze-Hard & ARC-1 & ARC-2 \\
\midrule
\multicolumn{6}{l}{\textit{Large language models}} \\
DeepSeek R1           & 671B & 0.0 & 0.0 & 15.8 & 1.3 \\
Claude 3.7 Sonnet      & --   & 0.0 & 0.0 & 28.6 & 0.7 \\
o3-mini-high          & --   & 0.0 & 0.0 & 34.5 & 3.0 \\
Gemini 2.5 Pro (32K)  & --   & --  & --  & 37.0 & 4.9 \\
\midrule
\multicolumn{6}{l}{\textit{Reproduced by us}} \\
TRM$^\dagger$          & 7M  & 73.6 & --   & --   & 4.6 \\
\midrule
\multicolumn{6}{l}{\textit{Recursive reasoning models (reported)}} \\
HRM                   & 27M & 55.0 & 74.5 & 40.3 & 5.0 \\
Attractor Model       & 27M & 91.4 & \textbf{93.1} & -- & -- \\
URM                   & 14M & 77.6 & --
                      & \textbf{$\geq$53.8}$^\ddagger$
                      & \textbf{$\geq$16.0}$^\ddagger$ \\
TRM                   & 7M  & 74.7 & 85.3 & 44.6 & 7.8 \\
EqR                   & 7M  & 93.0 & --   & --   & --  \\
Attractor Model       & 7M  & 54.3 & 46.7 & --   & --  \\
FPRM                  & 7M  & 94.2 & 87.0 & 47.5 & 6.2 \\
\rowcolor{ourblue}
\methodshort{} (ours) & 7M
                      & \textbf{\underline{97.9}}
                      & \underline{87.4}
                      & \underline{49.1}
                      & \underline{13.6} \\
\bottomrule
\end{tabular}
\end{table}

\subsection{Test-time depth and compute efficiency}
\label{sec:scaling}

We next evaluate how \methodshort{} translates recurrent computation into reasoning accuracy.
We conduct controlled evaluations on a fixed sample of Sudoku-Extreme test puzzles across varying layer budgets without retraining.
With two Transformer layers per shared block, executing $S$ supervision segments, $H$ outer cycles, and $L$ inner updates corresponds to $2SH(L+1)$ layer evaluations.

\begin{figure}[!tbp]
\centering
\includegraphics[width=\linewidth]{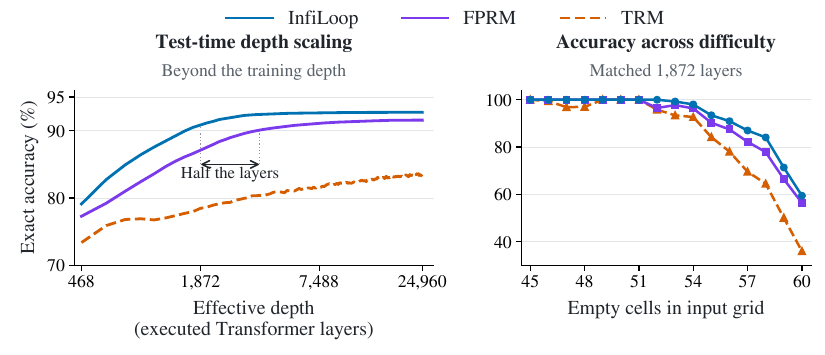}
\caption{\textbf{Test-time scaling and accuracy across puzzle difficulty.}
\textbf{Left:} Exact accuracy against executed Transformer layers (logarithmic axis), with all three models evaluated to 24,960 layers.
\methodshort{} reaches 90.9\% at 1,872 layers, exceeding FPRM's 90.1\% at 3,744 layers with half the layer evaluations.
\textbf{Right:} Exact accuracy by empty-cell count at matched 1,872 layers.
}
\label{fig:scaling_difficulty}
\label{fig:efflayers}
\label{fig:difficulty}
\end{figure}

\paragraph{Comparison with FPRM under test-time scaling.}
Figure~\ref{fig:efflayers} (left) compares \methodshort{} against the released FPRM checkpoint and retrained TRM across executed layers up to 24,960.
At 1,872 layer evaluations, \methodshort{} achieves 90.9\% exact accuracy overall.
This exceeds FPRM's 90.1\% accuracy at 3,744 layer evaluations, cutting the required computation in half.
When matched at 3,744 layer evaluations, \methodshort{} reaches 92.4\% overall, maintaining a steady advantage over FPRM at 90.1\%.
Furthermore, the accuracy that \methodshort{} achieves at 936 layer evaluations (86.5\%) is never matched by TRM even when evaluated at 24,960 layers (83.65\%), more than 26 times deeper.
At 24,960 layer evaluations, exact accuracy is 92.74\% for \methodshort{}, 91.63\% for FPRM, and 83.65\% for TRM.

\paragraph{Accuracy across puzzle difficulties.}
Figure~\ref{fig:difficulty} (right) evaluates accuracy as a function of the number of empty cells in the input grid, an established proxy for Sudoku difficulty.
We evaluate all difficulty levels from 45 to 60 empty cells on the 32,768 test puzzles at matched 1,872 layer evaluations.
For easy puzzles with 45 to 51 empty cells, both \methodshort{} and FPRM achieve 100.0\% exact accuracy.
From 52 empty cells onward, the margin of \methodshort{} over FPRM stays between 1.5 and 6.2 percentage points, while the margin over TRM rises from 4.2 percentage points at 52 empty cells to 23.4 at 60.
At 52 empty cells, \methodshort{} attains 100.0\% accuracy against 96.5\% for FPRM (+3.5 percentage points) and 95.8\% for TRM (+4.2 percentage points).
At 55 empty cells, \methodshort{} scores 93.4\% against 90.3\% for FPRM (+3.2 percentage points) and 84.2\% for TRM (+9.3 percentage points).
At 58 empty cells, the margin over FPRM peaks at $+6.2$ percentage points (84.1\% versus 77.9\%), while the margin over TRM widens to $+19.6$ percentage points.
At the most difficult count of 60 empty cells, \methodshort{} maintains 59.5\% accuracy against 56.5\% for FPRM (+3.0 percentage points) and 36.0\% for TRM (+23.4 percentage points).
Learned state reuse shields intermediate deductions precisely when search paths are deep and complex.

\subsection{What the connection learns to carry}
\label{sec:learned_step}

\begin{figure}[!tbp]
\centering
\includegraphics[width=\linewidth]{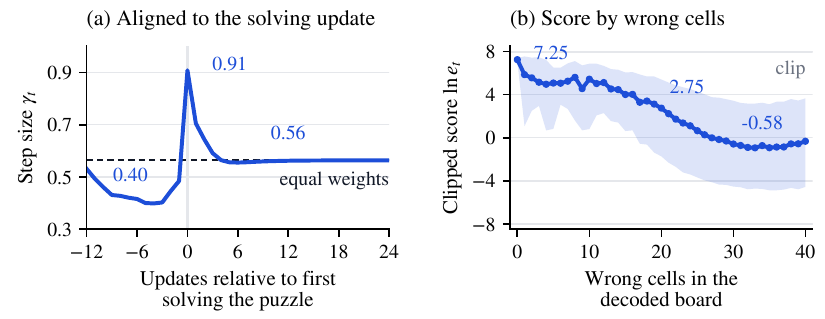}
\caption{\textbf{Learned adaptive updates reflect solution milestones and board correctness.} 
\textbf{(a)} Mean adaptive step size $\gamma_t$ aligned to the update at which the puzzle is first solved (for puzzles solved at update 13 or later). The step size remains suppressed ($\gamma_t \approx 0.40$) during search, peaks at $0.91$ when the puzzle is completed, and settles to the equal-weight baseline $1 - \beta = 0.56$ thereafter. 
\textbf{(b)} Median clipped content score $\ln e_t=\operatorname{clip}(s_t,-8,8)$ for block outputs produced after update 12, plotted against the number of incorrect cells in the decoded board. The shaded band denotes the interquartile range. High weights are assigned exclusively to valid configurations without explicit correctness supervision.}
\label{fig:step_size_dynamics}
\label{fig:learned_step}
\end{figure}

Section~\ref{sec:diagnosis} demonstrated that carry-last destabilizes solutions because later iterations overwrite earlier correct states.
\Eqref{eq:step_size} reveals how \methodshort{} regulates state propagation through the adaptive step size $\gamma_t = e_t / (\beta D_{t-1} + e_t)$.
If the scorer assigned equal weight to every block output, $\gamma_t$ would converge to $1-\beta = 0.56$.
Any divergence from 0.56 reflects learned content preferences.
Figure~\ref{fig:learned_step} analyzes how the step size and write scores behave across test puzzles.

\paragraph{The step size surges when the solution appears.}
Figure~\ref{fig:learned_step}a aligns puzzles solved at update 13 or later to their solving update.
Prior to finding the solution, the mean step size remains suppressed at approximately 0.40, below the equal-weight baseline of 0.56.
At the exact update that completes the puzzle, the step size surges to 0.91, giving dominant weight to the newly found solution.
Within six subsequent updates, it returns to the baseline value of 0.56.
This surge is strictly contingent on finding the solution; for puzzles that remain unsolved, the step size stays suppressed across all iterations.

\paragraph{The content score reflects solution correctness.}
The scorer receives no external hints or task labels during reasoning.
Figure~\ref{fig:learned_step}b plots the clipped content score $\ln e_t = \operatorname{clip}(s_t, -8, 8)$ for outputs after update 12 against the number of wrong cells in the decoded board.
The candidate states that decoding to a fully correct board achieves a median score of 7.25, close to the upper clipping bound of 8.0.
The assigned score drops monotonically as prediction errors increase: the median score falls to 2.75 at 20 wrong cells, and drops below zero to $-0.58$ at 33 wrong cells.
Valid candidate states receive high write weights and dominate the recurrent summary.
Candidate states with errors receive lower weights, limiting their influence on the reasoning trajectory.
Appendix~\ref{app:content_ablation} shows that removing the content score, during training or at test time, lowers accuracy by 6 to 7 percentage points.
\begin{figure}[!tbp]
\centering
\includegraphics{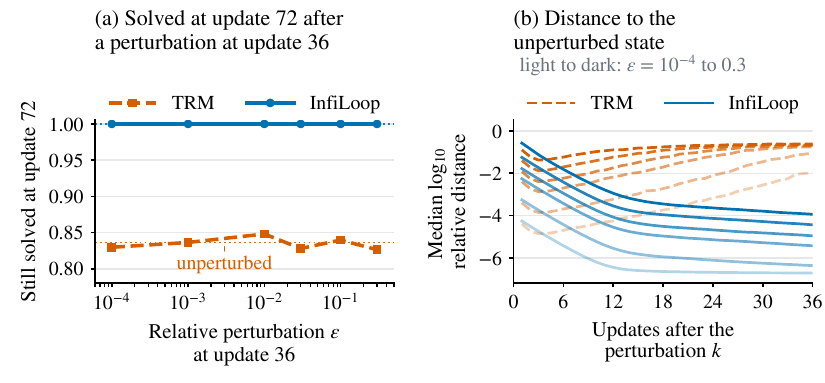}
\caption{\textbf{State stability and solution retention under recurrent state perturbation.} 
At outer update 36, a relative perturbation $\epsilon \in [10^{-4}, 0.3]$ is injected into the recurrent state, followed by 36 additional updates. 
\textbf{(a)} Proportion of puzzles solved at update 36 that remain solved at update 72 as a function of perturbation magnitude $\epsilon$. The horizontal dashed line marks the unperturbed baseline. 
\textbf{(b)} Evolution of the median relative Euclidean distance between the perturbed and unperturbed states, $\log_{10}(\|h'_{36+k} - h_{36+k}\| / \|h_{36+k}\|)$, across subsequent updates $k$. Different line shades denote perturbation magnitudes $\epsilon$. While perturbations in TRM quickly diverge to $O(1)$ relative distance, InfiLoop constrains trajectory drift and retains all solved states.}
\label{fig:perturbation_stability}
\label{fig:perturbation}
\end{figure}

\paragraph{State stability under recurrent perturbation.}
Because candidate states with constraint violations receive small content weights $e_t$, the recurrence limits the influence of noisy updates on the accumulated state.
To test whether this mechanism stabilizes the trajectory against perturbations, we inject a relative perturbation $\epsilon \in [10^{-4}, 0.3]$ into the recurrent state at update 36 and unroll the model for another 36 updates.
As shown in Figure~\ref{fig:perturbation}a, \methodshort{} retains 100\% of its solved puzzles at update 72 across all tested perturbation scales, while TRM loses 15--20\% of solutions even without perturbation.
Figure~\ref{fig:perturbation}b tracks the relative Euclidean distance to the unperturbed trajectory over subsequent updates $k$.
For TRM, even a minimal perturbation of $\epsilon = 10^{-4}$ causes the trajectory to diverge to a relative distance of order one within 12 updates.
In contrast, \methodshort{} keeps the displacement bounded throughout subsequent iterations, showing that the learned aggregation prevents chaotic drift in the recurrent state.

\subsection{Connection placement and decay initialization}
\label{sec:ablation}

We evaluate the design choices underlying \methodshort{}: the placement of the learned connection across recurrent loops, and the initialization of the temporal decay $\beta_0$.
Table~\ref{tab:abl_axis} summarizes both ablations on 7,680 puzzles sampled uniformly at random from the Sudoku-Extreme test set.

\paragraph{Applying the connection to both axes.}
The nested reasoner maintains two distinct recurrent trajectories: an inner loop for iterative candidate refinement, and an outer loop for context updates.
Table~\ref{tab:abl_axis} compares applying \methodshort{} to the inner loop alone, the outer loop alone, and both loops simultaneously.
The dual-axis configuration achieves the highest accuracy across all evaluated budgets.
At the training depth of $H3\,L6$, dual-axis aggregation reaches 82.1\% accuracy, compared with an inner connection alone at 72.6\% and an outer connection alone at 75.1\%.
This margin widens as the reasoning budget expands to $H32\,L6$, reaching 89.7\% against 79.4\% and 80.0\%.
Both recurrent trajectories benefit from learned state reuse.

\begin{table}[!tbp]
\centering
\caption{\textbf{Component and initialization ablations on Sudoku-Extreme.}
Exact solution accuracy (\%) on 7,680 test puzzles.
Top: placement of the connection across recurrent axes at training depth $H3\,L6$ and three larger budgets.
Bottom: temporal decay on the two-axis model. The left group fixes $\beta$ during training. The right group learns $\beta$ from the initialization $\beta_0$.}
\label{tab:abl_axis}
\small
\setlength{\tabcolsep}{4pt}
\renewcommand{\arraystretch}{1.15}
\begin{tabular*}{\linewidth}{@{\extracolsep{\fill}}l rrrr@{}}
\toprule
\multicolumn{5}{@{}l}{\textbf{Connection placement across budgets}} \\
\cmidrule{1-5}
Configuration & $H3\,L6$ & $H3\,L12$ & $H3\,L96$ & $H32\,L6$ \\
\midrule
Inner axis alone & 72.6 & 75.2 & 77.2 & 79.4 \\
Outer axis alone & 75.1 & 76.3 & 76.2 & 80.0 \\
\textbf{Both axes (ours)} & \textbf{82.1} & \textbf{85.3} & \textbf{87.4} & \textbf{89.7} \\
\bottomrule
\end{tabular*}
\par\vspace{5pt}
\begin{tabular*}{\linewidth}{@{\extracolsep{\fill}}l rrr rrr@{}}
\toprule
\textbf{Temporal decay} & \multicolumn{3}{c}{Fixed $\beta$} & \multicolumn{3}{c}{Learned $\beta_0$} \\
\cmidrule(lr){2-4}\cmidrule(l){5-7}
Budget & Reverse & Uniform & Forward & $0.10$ & $0.90$ & $0.55$ (ours) \\
& $1.82$ & $1$ & $0.55$ & & & \\
\midrule
$H3\,L6$ & 2.0 & 38.4 & 74.0 & 66.3 & 63.3 & \textbf{82.1} \\
$H32\,L6$ & 2.0 & 40.2 & 77.9 & 72.2 & 66.0 & \textbf{89.7} \\
\bottomrule
\end{tabular*}
\end{table}

\paragraph{Forward decay over the loop trajectory.}
The decay $\beta=e^{-\lambda}$ sets how fast older block outputs lose weight in the aggregated state.
Forward decay uses $\beta<1$, so older outputs weigh less.
Uniform weights use $\beta=1$, which is the Full Attention Residuals case $\lambda=0$.
Reverse decay uses $\beta>1$, so older outputs weigh more.
We first fix $\beta$ during training for the two-axis model.
Reverse decay with $\beta=1.82$ solves only 2.0\% of the puzzles at $H3\,L6$.
Uniform weights reach 38.4\%, and forward decay with $\beta=0.55$ reaches 74.0\%.
The gap grows with the budget: at $H32\,L6$, forward decay reaches 77.9\%, uniform weights reach 40.2\%, and reverse decay stays at 2.0\%.
Learning $\beta$ from $\beta_0=0.55$ gives the best accuracy, 82.1\% at $H3\,L6$ and 89.7\% at $H32\,L6$.
The initialization also matters: $\beta_0=0.10$ reaches 66.3\% and $\beta_0=0.90$ reaches 63.3\% at $H3\,L6$.
We therefore learn a forward decay from $\beta_0=0.55$ on all benchmarks.

\section{Conclusion}
We discuss related work in Appendix~\ref{sec:related}.
In this paper, we identify state replacement under the carry-last rule as a primary bottleneck in deep looped Transformers: repeatedly replacing the recurrent state with the latest block output overwrites correct intermediate deductions and destabilizes solved configurations. To address this, we introduce InfiLoop, a loop-native residual connection that combines content-based pseudo-query scoring with exponential temporal decay across iterations.
Its exact recurrence needs only $O(d)$ persistent memory per token and keeps the aggregated state within the convex hull of the candidate representations.
On Sudoku-Extreme, Maze-Hard, ARC-AGI-1, and ARC-AGI-2, InfiLoop outperforms carry-last and fixed-point damping baselines, and it reaches their accuracy with fewer layers under test-time depth scaling.
Furthermore, analysis of the learned connection reveals that it operates as an adaptive residual update: the model suppresses proposals with constraint violations, increases update magnitude upon reaching a valid solution, and resists state drift under perturbations.
These findings show that learning how to selectively aggregate recurrent states is an effective mechanism for scaling test-time reasoning depth.

\bibliographystyle{report}
\bibliography{iclr2027_conference}

\newpage
\appendix
\section{Compute and Persistent-State Accounting}
\label{app:resources}

We quantify the computation and persistent state of \methodshort{}.
It stores 582.8~KiB of persistent state per puzzle, independent of the loop budget.

\paragraph{Layer evaluations.}
Each outer cycle applies the shared block $L$ times to the inner state and once to the outer state.
With $S$ supervision segments, $H$ outer cycles per segment, and $K$ layers per block, the number of Transformer layer evaluations is
\begin{equation}
C(H,L)=S H(L+1)K.
\label{eq:compute}
\end{equation}
Training uses $S{=}12$, $H{=}3$, $L{=}6$, and $K{=}2$, which gives 504 layer evaluations.
The depth study in Section~\ref{sec:scaling} fixes $H{=}6$ and $L{=}12$, so each segment executes 156 layers, and varies $S$ from 1 to 160, which gives 156 to 24,960 layer evaluations.
This count omits the connection and the prediction heads, which we add below.
The embedding is a table lookup and adds no multiply-accumulates.
FPRM also uses a two-layer block with width 512 and the same sequence length.
Layer evaluations therefore match the dominant Transformer arithmetic across the two models.
FPRM's solver operations are outside the InfiLoop FLOP accounting below.
These counts do not compare measured runtime.

\paragraph{Profiling setup and FLOP convention.}
We profile the InfiLoop Sudoku checkpoint at batch size one under the 12-segment inference protocol.
Each sequence has 81 board cells and 16 prefix positions, so the shared block processes $n{=}97$ positions.
The SwiGLU intermediate width is $m{=}1{,}536$: the implementation takes two thirds of the expansion $4d$ and rounds it up to a multiple of 256.
We count two FLOPs per multiply-accumulate in linear projections, attention matrix products, the depthwise convolution, and the content scores of the connection.
Normalization, activation functions, rotary embeddings \citep{su2024roformer}, other elementwise arithmetic, and data movement are not counted.

\paragraph{Counted FLOPs.}
Let $d$ be the hidden width, $g$ the number of board cells, and $k_c$ the convolution kernel width.
One application of the shared block costs
\begin{equation}
F_{\mathrm{block}}=K(8nd^2+4n^2d+6ndm)+2gdk_c^2,
\end{equation}
where the convolution applies once per block to the $g$ board cells and the attention and feed-forward projections apply to all $n$ positions.
For $n{=}97$, $d{=}512$, $m{=}1536$, $K{=}2$, $g{=}81$, $k_c{=}3$, this is 1.36154~GFLOPs per block application in \methodshort{}.
The prediction head has $V{=}11$ outputs per position.
The halting head of the adaptive computation procedure has two outputs and reads one position.
With $T{=}SH(L+1)$ block applications, the counted total is
\begin{equation}
F_{\mathrm{InfiLoop}} = T F_{\mathrm{block}}+S(2ndV+4d)+2nd(T+I),\qquad I=SH+2S.
\end{equation}
The term $2ndT$ is the content score of every block output.
The term $2ndI$ counts the content scores of initialization writes.
These comprise one write per outer cycle and two writes per segment in the carry reset.
The carry reset replaces the recurrent state of a puzzle that halted in the previous segment with the learned initial state.
The carry reset is scored for every puzzle, because the implementation builds both the reset and the retained state before selecting one.
We count the operators of one segment in the \methodshort{} checkpoint and multiply by $S{=}12$.
This count matches the formula at four profiled depths: 0.34, 0.92, 2.55, and 25.6 TFLOPs per puzzle at $H{=}3,L{=}6$, $H{=}8,L{=}6$, $H{=}12,L{=}12$, and $H{=}32,L{=}48$.
At the depth-study setting $H{=}6,L{=}12$, the formula gives 1.27 TFLOPs at 1,872 layers and 17.0 TFLOPs at 24,960 layers.

The content scores are the only operations of the connection in the totals above.
The remaining arithmetic of the connection is elementwise.
We charge each arithmetic primitive or unary function once and exclude comparisons, casts, indexing, and memory traffic.
Under this convention, the remaining arithmetic adds $T[n(7d+6)+1]+In(4d+4)$ scalar operations.

\paragraph{Persistent state.}
The shared block attends bidirectionally over the whole input and recomputes keys and values at every application.
InfiLoop keeps no KV cache between iterations.
What persists between block applications is the recurrent carry.
At batch size one, the two BF16 carries occupy $2nd\cdot2$ bytes, or 194.0~KiB.
A carry-last TRM keeps the same two carries.
\methodshort{} adds an FP32 numerator and denominator per position on each axis, $2n(d+1)\cdot4$ bytes or 388.8~KiB, for a total of 582.8~KiB.
None of these sizes depends on $H$, $L$, or $S$.
They scale linearly with batch size and sequence length.
Model weights, input tensors, and temporary operator workspaces are not part of this accounting.

\section{Related Work}
\label{sec:related}
\paragraph{Reasoning through recurrent depth.}
Universal Transformers apply one shared block repeatedly \citep{dehghani2019universal}.
Later work studied the computational capabilities of looped Transformers and their use for latent reasoning \citep{giannou2023looped,saunshi2025reasoning,geiping2025scaling}.
Looped Transformers have also been studied as learners of iterative algorithms \citep{yang2024looped} and for generalization to longer input sequences \citep{fan2025length}.
HRM and TRM build compact reasoners from nested latent updates \citep{wang2025hierarchical,jolicoeurmartineau2025less}.
URM studies architecture and optimization choices for recursive reasoning \citep{gao2025universal}.
Adaptive Computation Time, PonderNet, and Mixture-of-Recursions learn how much computation to spend \citep{graves2016adaptive,banino2021pondernet,bae2025mixture}.
\methodshort{} modifies the connection between iterations of a nested TRM-style reasoner.

\paragraph{Overthinking in recurrent networks.}
\citet{schwarzschild2021algorithm} showed that recurrent networks trained on easier problems can generalize to harder ones by taking more iterations at test time.
\citet{bansal2022overthinking} identified degradation under excessive iteration and addressed it with input recall and progressive training.
Their recall connection preserves access to the original problem instance, while \methodshort{} aggregates candidate states produced during reasoning.
Both address behavior beyond the training iteration budget, but act on different information in the recurrence.

\paragraph{Stabilizing deep and looped Transformers.}
A looped Transformer at many iterations is a very deep weight-shared network.
Section~\ref{sec:diagnosis} measures one consequence: carry-last TRM accuracy peaks and then falls with depth.
Normalization placement affects gradient stability at initialization \citep{xiong2020layer}, while depth-dependent normalization scaling controls activation growth \citep{sun2025curse}.
ReZero uses a zero-initialized residual gate \citep{bachlechner2021rezero}, LayerScale learns channel-wise residual scaling \citep{touvron2021layerscale}, and DeepNet combines residual scaling with a tailored initialization \citep{wang2024deepnet}.
Residual scaling for looped Transformers controls residual growth under weight sharing \citep{wang2026residual}.
Deep Equilibrium Models formulate weight-tied networks through fixed-point root finding and implicit differentiation \citep{bai2019equilibrium}.
FPRM combines pre-norm, learned residual scaling, and damped fixed-point updates, with convergence-based halting \citep{movahedi2026fixed}.
Attractor Models and equilibrium reasoners use fixed-point formulations for latent reasoning \citep{feinashley2026solve,huang2026equilibrium}.
At a fixed point the update size is zero.
These approaches stabilize recurrent computation through normalization, update scaling, or fixed-point formulations.
\methodshort{} instead learns from the task loss which earlier outputs the next state keeps and by how much.

\paragraph{Learned residual connections.}
ResNet propagates features through identity skip connections \citep{he2016resnet}, while DenseNet concatenates features from preceding layers \citep{huang2017densenet}.
DenseFormer learns weights for averaging current and earlier layer outputs \citep{pagliardini2024denseformer}.
Hyper-Connections learn how to mix features across depth and parallel residual streams \citep{zhu2025hyperconnections}.
These works establish learned cross-layer information flow as a broader design space.
\methodshort{} specializes this idea to a shared recurrent block: content-dependent weights and exponential age decay admit a running numerator and denominator whose size is independent of the number of iterations.

\paragraph{Gated recurrent updates.}
The interpolation in \Eqref{eq:step_size} has the same form as an update gate in a GRU \citep{cho2014encoderdecoder} or the coupled transform and carry gates in a Highway Network \citep{srivastava2015highway}.
The distinction is the gate parameterization: $\gamma_t$ is the ratio of the new content weight to the sum of that weight and the decayed historical mass, rather than the output of a separate gating network.
One shared content scorer and one decay parameter add $d+1$ parameters per recurrence axis.
This parameterization gives the exact normalized history aggregation in \Eqref{eq:attn_res}.

\paragraph{Relation to Attention Residuals.}
Full Attention Residuals learns input-dependent weights over the outputs of earlier layers \citep{kimi2026attention}.
Each layer has its own pseudo-query and attends to the stored outputs of earlier layers, so memory grows with depth.
A blockwise variant stores block-level summaries to reduce this history.
A looped model has no fixed number of iterations, and a loop-native connection therefore cannot store the full activation history.
\methodshort{} scores each output once, at its own iteration, with one scorer shared by every iteration.
A learned exponential decay then ages this weight.
The connection keeps only a running weighted sum and a running normalization mass, so its memory does not grow with the loop count and the same rule applies at any depth.
Retention networks also combine exponential age weighting with a recurrent form \citep{sun2023retentive}, and ALiBi demonstrates input-length extrapolation with linear relative-position biases \citep{press2022alibi}.
In \methodshort{}, the recurrence runs over reasoning iterations and yields the next recurrent state.

\section{Qualitative analysis on ARC-AGI-2.}
\begin{figure}[t]
\centering
\includegraphics[width=\linewidth]{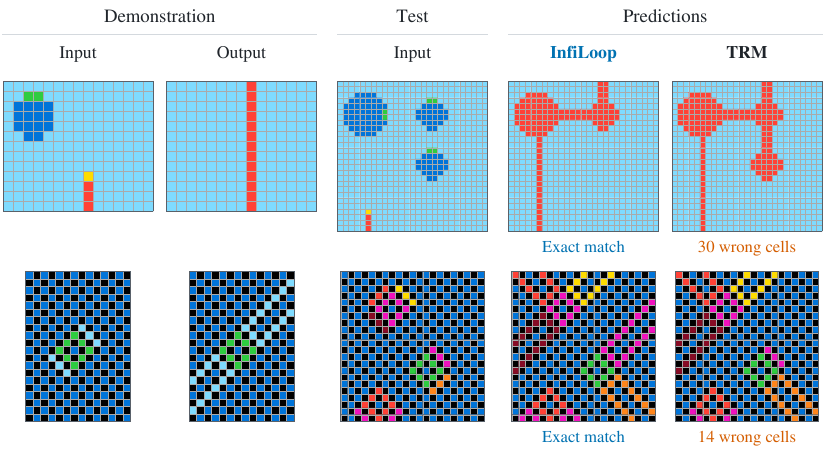}
\caption{\textbf{Qualitative comparison on ARC-AGI-2.}
Each row shows a demonstration input--output pair, a test input, and the predictions of \methodshort{} and TRM.
The top and bottom rows correspond to tasks \texttt{5961cc34} and \texttt{80a900e0}, respectively.
As in the pass@2 evaluation, we show the best of each model's two submitted attempts.
\methodshort{} matches the target exactly in both cases; TRM makes 30 and 14 cell errors, respectively.}
\label{fig:arc2_cases}
\end{figure}

Figure~\ref{fig:arc2_cases} visualizes the qualitative predictions of \methodshort{} and TRM on two held-out ARC-AGI-2 evaluation tasks.
In Task \texttt{5961cc34}, which demands geometric expansion of spatial components, \methodshort{} produces the exact target structure, while TRM produces 30 erroneous cells with distorted boundaries.
In Task \texttt{80a900e0}, the model must extend colored diagonal rays across an alternating checkerboard background.
\methodshort{} preserves the underlying background regularity while drawing sharp diagonal rays.
TRM introduces 14 corrupted cells that break the checkerboard pattern and fragment the rays.
These qualitative results demonstrate that learned state reuse provides stability across complex spatial rules where carry-last propagation degrades structural coherence.

\section{Removing the Content Score}
\label{app:content_ablation}

Section~\ref{sec:learned_step} shows that the content score tracks board correctness.
We test whether this score also matters for accuracy by removing it in two ways.
Setting $q=0$ gives every candidate the same score $s_i=0$, and \Eqref{eq:step_size} then reduces the connection to an exponential moving average whose step size approaches $1-\beta$.
All results in this section use the Sudoku-Extreme test puzzles of Section~\ref{sec:scaling}, evaluated at $H6\,L12$ with $S{=}12$, or 1,872 layers.
Table~\ref{tab:content_ablation} summarizes them.
\begin{table}[h]
\centering
\caption{Exact accuracy (\%) on Sudoku-Extreme at $H6\,L12$, $S{=}12$, or 1,872 layers.}
\label{tab:content_ablation}
\small
\begin{tabular}{cccc}
\toprule
\methodshort{} & $q=0$ in training & $q=0$ at test time & $\beta=0$ at test time \\
\midrule
90.9 & 84.1 & 84.8 & 1.5 \\
\bottomrule
\end{tabular}
\end{table}
\paragraph{Removing the score during training.}
We train a second model with the same recipe, seed, and number of training steps, and with $q$ fixed to zero on both axes.
It reaches 84.1\% against 90.9\% for \methodshort{}, so the content score is worth 6.8 percentage points when the model is trained with it.

\paragraph{Removing the score at test time.}
We also set $q$ to zero in the trained \methodshort{} checkpoint without retraining, and accuracy falls from 90.9\% to 84.8\%.
Setting $\beta$ to zero instead turns the connection into carry-last, and accuracy falls to 1.5\%, because the block weights are trained to operate on the aggregated state rather than on raw block outputs.

Both ways of removing the score lower accuracy by 6 to 7 percentage points.
The gain of \methodshort{} therefore does not come from damping alone: a fixed exponential average also damps the updates but lacks the content score, and it is less accurate.

\end{document}